\documentclass[11pt]{article}
\ifdefined\pdfminorversion\pdfminorversion=7\fi 

\PassOptionsToPackage{hyperfootnotes=false}{hyperref}
\usepackage[preprint]{acl}

\usepackage{times}
\usepackage[T1]{fontenc}
\usepackage[utf8]{inputenc}

\newcommand{\creditreferenceproviderphrase}{a Claude-family external LLM}
\newcommand{\sftteacherproviderphrase}{a stronger Claude-family agentic teacher model}
\newcommand{\simulatorproviderphrase}{the same Claude-family instruction-following simulator}
\newcommand{\venuecompacttablefont}{\small}

\usepackage{microtype}
\usepackage{inconsolata}
\usepackage{graphicx}
\usepackage{booktabs}     
\usepackage{amsmath}
\usepackage{amssymb}
\usepackage{enumitem}     
\usepackage{placeins}     
\usepackage{url}
\usepackage{xspace}

\graphicspath{{figures/}}

\newcommand{\method}{SGCD\xspace}

\providecommand{\creditreferenceproviderphrase}{a training-time external LLM}
\providecommand{\sftteacherproviderphrase}{a stronger external agentic teacher model}
\providecommand{\simulatorproviderphrase}{the same external instruction-following simulator}

\providecommand{\suppref}[1]{Appendix~\ref{#1}}
\providecommand{\supptabref}[1]{Appendix Table~\ref{#1}}

\title{Keep Policy Gradient in Charge: Sibling-Guided Credit Distillation
       for Long-Horizon Tool-Use Agents
       \thanks{Work done at Amazon Web Services. Correspondence:
       \texttt{tianyd@amazon.com}.}}

\author{
  Tianyu Ding \\
  Amazon Web Services \\
  \texttt{tianyd@amazon.com} \\\And
  Jianhong Xin \\
  Amazon Web Services \\
  \texttt{xijianho@amazon.com} \\\And
  Juan Pablo De la Cruz Weinstein \\
  Amazon Web Services \\
  \texttt{jcruam@amazon.com} \\
}

\begin{document}
\maketitle

\begin{abstract}
Long-horizon tool-use RL learns from outcome verification, but trajectory-level
advantages are broadcast over many reasoning, API, and answer tokens. Direct
self-distillation seems to offer a denser signal, yet in the direct-SD recipes
we study it can \emph{collapse} tool use. We diagnose a plausible failure path:
teacher-matching pressure can rehearse behavior without knowing which actions
the verifier rewards, reinforcing useful skills and shortcuts together. We
introduce \textbf{Sibling-Guided Credit Distillation (SGCD)}, which
uses distillation for bounded credit weighting rather than as a competing actor
loss. Dynamic sampling yields mixed successful/failed sibling rollouts; an
external LLM summarizes their contrast into a training-only credit reference;
and detached teacher/student divergence reshapes policy-gradient token advantages. The
inference-time student sees no external LLM, sibling evidence, or oracle. On
AppWorld, SGCD reports higher held-out point estimates than Vanilla GRPO under
matched rollout-generation hyperparameters and evaluation settings:
TGC $42.9\to45.6$ on \texttt{test\_normal} and $24.7\to27.0$ on
\texttt{test\_challenge}. This is a recipe-level comparison: Vanilla GRPO
includes reference-KL, whereas SGCD omits it and uses detached divergence for
credit weighting. On $\tau^3$-airline, three independent SGCD training
runs give a directional higher held-out mean Pass$^1$ estimate,
$0.583\to0.615\pm0.006$,
against the Vanilla GRPO comparator, whose benchmark recipe includes
reference-KL while SGCD omits it.

\end{abstract}

\section{Introduction}
\label{sec:intro}

Recent multi-turn tool-agent work \citep{coevolve2026,multiturnguide2025} uses
group-relative policy-gradient (PG) methods such as GRPO
\citep{deepseekmath2024}. These methods can
learn from outcome verification alone, but their token-level credit assignment is
coarse: the same trajectory-level advantage is broadcast across long sequences
of reasoning, API calls, tool observations, and final answers. Self-distillation
(SD) offers an attractive complement: use the policy's own successful behavior,
or a teacher conditioned on privileged training-time information, to supply a
denser signal \citep{sdpo2026,opsd2026,skillsd2026,sdar2026}. The hard part is
making that signal help the policy gradient rather than fight it.

\paragraph{The hidden failure.} We first show, on the corrected $\tau^3$-airline
customer-service task set in the $\tau$-bench family
\citep{taubench2024,tau2bench2025,tau3bench2026}, that naive SD
\emph{can degrade} long-horizon tool-use. Over training, the model
increasingly solves easy information-only cases without using tools, while
state-changing action tasks collapse to zero success (\S\ref{sec:diagnosis}).
The aggregate score conceals the mechanism: the policy is not learning a better
agent; it is abandoning executable tool-use behavior the base or supervised
checkpoint already possessed.

\paragraph{Design implication: use distillation for credit, not as a competing loss.}
The contrast with a $\tau^3$ Vanilla GRPO control using the benchmark
reference-KL is informative: it preserves
tool-use and reaches held-out mean Pass$^1$ $0.583$, whereas SDPO reaches
$0.317$ and collapses on state-changing action tasks. Our subsequent
AppWorld probes led to the
same design constraint. Direct SD pressure can be live and non-inert, yet still
degrade action quality when the teacher target is misaligned with what the
verifier rewards; our diagnostic direct/uncapped SD ablation
falls below the reported Vanilla GRPO control. This evidence motivates
repurposing
teacher/student divergence as a \emph{credit-weighting signal}: it determines
where the broadcast policy-gradient advantage should receive bounded
token-level emphasis, while the
reward-grounded policy optimization remains the update driver. On AppWorld,
this detached signal replaces Vanilla GRPO's reference-KL operationally, but
affects the actor only through credit weights rather than an equivalent penalty.

\paragraph{The method.} We introduce \textbf{Sibling-Guided Credit Distillation
(SGCD)}. Dynamic sampling produces mixed groups of verified successful and failed
sibling rollouts for the same task; \creditreferenceproviderphrase{} summarizes their
contrast into a stepwise credit reference (what succeeded, which branches failed,
where failures deviated). The policy is scored twice on the rollout: once under
the ordinary student context and once under the credit reference as a stop-gradient
teacher. Dense top-$K$ reverse-KL and entropy convert the divergence into
clipped token-level advantage weights inside the policy-gradient surrogate
(\S\ref{sec:method}). No distillation loss touches the actor: distillation guides
credit, and policy gradient performs the update.

\paragraph{Training-time boundary.} The external LLM, sibling evidence, and
credit reference are training-only; the inference-time student sees the same clean
task prompt as the policy, with no oracle, sibling rollouts, external LLM, or credit
reference. Because the divergence signal is detached from the actor parameters
and only reshapes the verified advantage, SGCD removes the teacher-matching
actor-gradient path that can pull the model toward actions the clean deployment
prompt cannot reproduce.

\paragraph{Contributions.}
\begin{enumerate}
  \item A tool-use diagnosis: naive self-distillation can reinforce benchmark
    shortcuts, preserving easy information-only behavior while losing
    state-changing action capability (\S\ref{sec:diagnosis}).
  \item \textbf{SGCD}: a sibling-guided credit-distillation method that builds a
    stepwise credit reference from mixed successful/failed rollouts and uses
    dense teacher/student divergence to reweight, not replace, the PG update
    (\S\ref{sec:method}).
  \item An empirical study on AppWorld and $\tau^3$-airline: AppWorld carries the
    main positive held-out result, while $\tau^3$-airline provides a directional
    held-out comparison and a failure-mode diagnostic. Ablations give lower
    point estimates when the sibling credit reference or divergence-driven
    weighting is removed, and direct SD pressure falls below the reported Vanilla
    GRPO control (\S\ref{sec:experiments}).
\end{enumerate}

\section{Background and Related Work}
\label{sec:related}

\paragraph{Group-relative RL for tool-use agents.} Policy-gradient RL builds on
score-function estimators and clipped optimization
\citep{williams1992reinforce,sutton1999policy,ppo2017}. GRPO
\citep{deepseekmath2024} estimates advantages from rollout groups without a
value function, and DAPO \citep{dapo2025} adds dynamic sampling over
mixed-outcome groups; GRPO-style policy-gradient recipes are now practical
backbones for reasoning and tool-use RL \citep{coevolve2026,multiturnguide2025}.
SGCD keeps this backbone
intact, reshaping token credit inside the surrogate without replacing the reward
or training a value model.

\paragraph{Knowledge distillation and privileged information.} Distillation
transfers a teacher's predictive distribution to a student \citep{hinton2015};
privileged-information learning conditions training on signals absent at
inference \citep{vapnik2015lupi}, and on-policy distillation reduces exposure
mismatch via the student's own samples \citep{agarwal2024onpolicykd}. GATES
\citep{gates2026} applies a same-model teacher under privileged context for
trajectory-level self-distillation. SGCD instead uses the privileged comparison
only as detached credit weighting inside the policy gradient.

\paragraph{Self-distillation for RL.} Recent LLM post-training uses
self-distillation to replace scalar reward or augment group-relative RL. SDPO
\citep{sdpo2026}, OPSD \citep{opsd2026}, Skill-SD \citep{skillsd2026}, and SDAR
\citep{sdar2026} cover feedback-conditioned self-teachers, label-privileged
forward-KL/JSD variants (JSD; \citealp{lin1991jsd}), skill-summary teachers,
and gap-gated token-level SD auxiliaries; GEAR \citep{gear2026} is closest,
using reverse-KL and entropy for
adaptive token/segment weights. \supptabref{tab:sdfamily} contrasts these with
SGCD, whose distinguishing choices are mixed successful/failed siblings, a
training-only external-LLM credit reference with no inference-time dependency,
and dense reverse-KL/entropy used for bounded advantage weights rather than a
standalone SD auxiliary. These methods motivate SGCD's teacher-side evidence and
dense comparisons. Our negative result motivates a narrower role in the studied
long-horizon tool-use settings: transferring teacher-side information through
detached weights on verified advantages rather than through a teacher-matching
actor loss.

\paragraph{Self-distillation instability.} Prior work reports instability from
unreliable, instance-specific, or prefix-conditioned teacher guidance
\citep{sdar2026,manyfaces2026,revisitopd2026}. Separately, OPD success depends
on teacher--student thinking-pattern compatibility and transferable novelty
\citep{thunlpopd2026}. We add a tool-use instance: zero-tool collapse on $\tau^3$-airline
(\S\ref{sec:diagnosis}), where a locally plausible teacher action can be
unreproducible under the clean deployment prompt.

\section{The Failure Mode: Naive SD Can Collapse Tool Use}
\label{sec:diagnosis}

\paragraph{Setup.} We train Qwen3.5-4B on the corrected $\tau^3$-airline task
set in the $\tau$-bench family
\citep{taubench2024,tau2bench2025,tau3bench2026} and evaluate 20 held-out task
specifications on a paired grid of three evaluation seeds and four greedy trials
per seed (240 episodes). Gold actions define \emph{state-changing action} tasks
(booking/cancellation/update; 13), \emph{information-only} tasks (4),
\emph{transfer-only} tasks (1), and \emph{no-action} tasks (2).

\paragraph{Aggregate result.} Table~\ref{tab:tau3} reports the held-out
$\tau^3$ mean Pass$^1$ across recipes. The Vanilla GRPO comparator reaches $0.583$; SDPO reaches
$0.317$ at the reported checkpoint and scores $0\%$ on state-changing action
tasks at late steps. The aggregate hides this category structure.

\paragraph{Observed behavior: zero-tool collapse.} A per-step rollout audit shows that
SDPO shifts toward no-tool information-only successes: the no-tool share rises
sharply, while mean tools per success falls $2.5\to0.0$. The Vanilla GRPO control
stays at $0\%$ zero-tool. Thus SDPO is not merely
trading off categories; it preserves easy information-only answers while
abandoning executable action behavior, as Table~\ref{tab:tau3} shows.

\paragraph{A plausible direct-SD failure path.} Stop-gradient prevents
teacher-parameter updates, but a teacher-matching target distribution can still
pull the student. If the teacher favors ``skip the tool, answer directly'' or an
action plan unsupported by the clean prompt, a KL/JSD auxiliary can optimize
that target and steer the actor away from what the verifier rewards. The
behavioral audit establishes collapse; it does not by itself identify this
teacher-target pathway as the unique cause.

\paragraph{Design implication.} This diagnosis motivates keeping the
policy-gradient objective as the
update driver. Distillation is useful for the narrower question: \emph{which parts
of a long trajectory deserve extra emphasis relative to the broadcast verified
advantage?} SGCD (\S\ref{sec:method}) uses teacher/student divergence for credit
weighting in a policy-gradient update, not as an independent actor-gradient loss.

\paragraph{Relation to known SD instability.} Prior work reports SD instability
on math reasoning and dialogue (\S\ref{sec:related}); zero-tool collapse is the
corresponding \emph{tool-use} instance and motivates keeping the reward-grounded
update primary.


\begin{figure*}[!t]
  \centering
  \includegraphics[width=\textwidth]{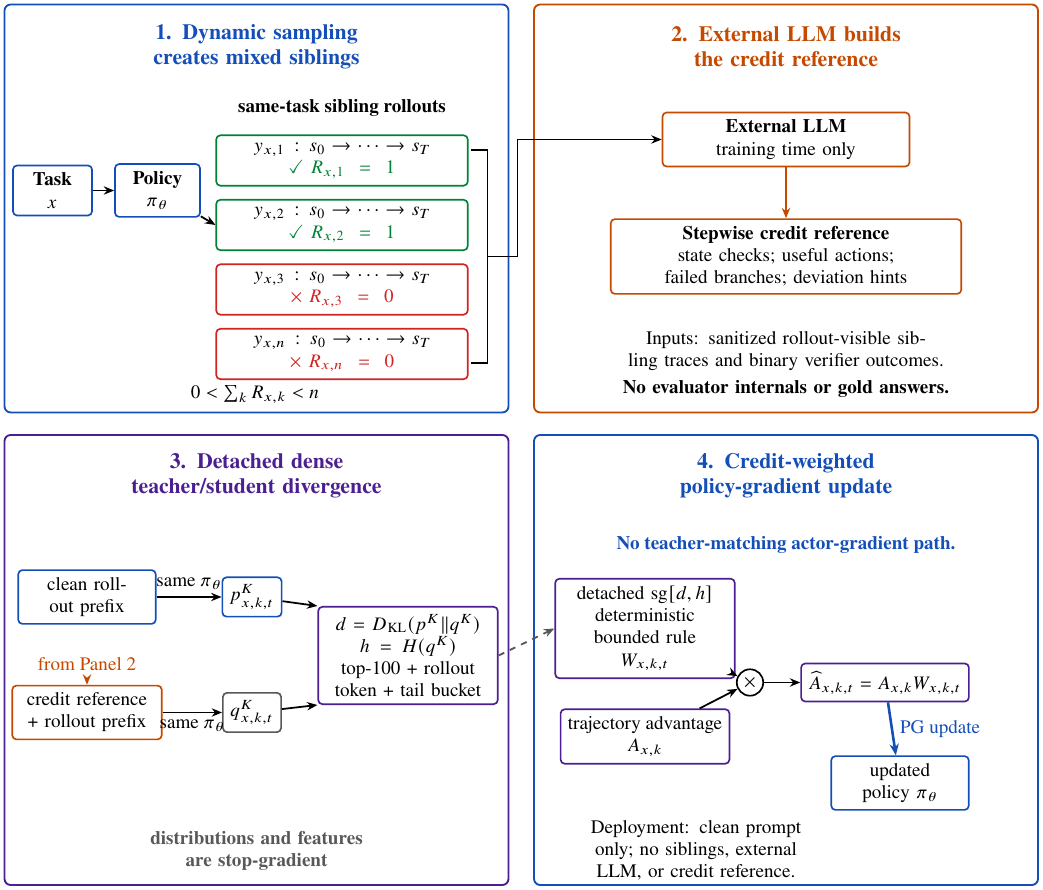}
  \caption{\method overview. Dynamic sampling creates mixed sibling rollouts;
    an external LLM summarizes their contrast into a training-only stepwise
    credit reference; detached dense teacher/student divergence produces
    credit features; and bounded credit weights reweight the policy-gradient
    advantage. The inference-time student receives only the clean task prompt.
    The schematic shows only SGCD-specific paths: the detached divergence
    pathway contributes no teacher-matching actor gradient. The reference-KL
    used by the Vanilla GRPO comparators is outside this SGCD schematic; SGCD
    uses no reference-KL.}
  \label{fig:method}
\end{figure*}

\section{Sibling-Guided Credit Distillation}
\label{sec:method}

\subsection{Overview}
\method (Figure~\ref{fig:method}) augments a group-relative policy-gradient
backbone with a distillation-derived credit signal. Dynamic sampling produces
mixed successful/failed sibling rollouts; an external LLM turns them into a
stepwise credit reference; the policy is scored under clean and credit-reference
contexts; and detached teacher/student divergence reshapes token advantages
inside the policy-gradient update. The inference-time model is the ordinary student policy: no external LLM,
credit reference, or sibling evidence appears at inference.

\subsection{Mixed sibling evidence}
\label{sec:siblings}
For a task $x$, the rollout worker samples a sibling group
$Y_x=\{y_{x,1},\ldots,y_{x,n}\}$ under the current policy, with verifier outcome
$R_{x,k}\in\{0,1\}$. A group is useful when it is \emph{mixed}:
\begin{equation}
  0 < \sum_{k=1}^{n} R_{x,k} < n .
\end{equation}
Mixed groups contain verified successful routes and failed alternatives under
the same task prompt, making dynamic sampling SGCD's sibling evidence source.
Following DAPO-style dynamic sampling \citep{dapo2025}, all-success and
all-failure candidate groups are discarded and replacement task groups are
sampled. The reported recipe refills toward eight complete mixed groups for at
most ten generation batches; if the target remains unmet, it skips that
optimizer update and continues training. Homogeneous groups never enter an SGCD
policy update.

\paragraph{Credit-reference construction.} For each mixed group, a strong
external LLM converts policy-visible sibling traces plus binary verifier
outcomes into a training-only stepwise credit reference $z_x$: reusable state
checks, useful actions, failed
branches, and likely deviation points. We rebuild this reference from fresh
on-policy siblings rather than a persistent skill bank (prompt details:
\suppref{sec:appendix}).

\paragraph{Leakage boundary.} The credit reference is never shown to the
inference-time student. During construction we retain only rollout-visible
traces and the binary verifier outcome; all other evaluator-private fields,
gold actions, sensitive values, and answer-bearing literals are removed. The
reference therefore abstracts reusable credit evidence rather than hidden
answers.

\subsection{Dense teacher/student divergence}
\label{sec:dense}
For a rollout token position $t$, let $p_{x,k,t}^{K}$ be the current policy
scored on the clean rollout prefix, and let $q_{x,k,t}^{K}$ use the same policy
weights on the same observed prefix with credit reference $z_x$ inserted as
teacher context. Both distributions are projected onto the same local support.
The reported runs use $K{=}100$. The support
contains the teacher top-$K$ tokens, the observed rollout token, and a tail
bucket that preserves remaining mass. We primarily use reverse-KL,
\begin{equation}
  d_{x,k,t}
  = D_{\mathrm{KL}}\!\left(
      p_{x,k,t}^{K} \;\middle\|\; \mathrm{sg}[q_{x,k,t}^{K}]
    \right),
  \label{eq:rkl}
\end{equation}
The reported SGCD results use reverse-KL. Equation~\ref{eq:rkl} stops gradients
through $q$; the complete divergence/entropy feature is detached from the actor
through $u$ below. Thus the divergence and teacher entropy are consumed as
credit features, not as a direct actor-gradient distillation loss.

\subsection{Divergence-driven credit weighting}
\label{sec:segments}
Long-horizon traces contain large stretches of boilerplate reasoning, setup, and
tool output. SGCD uses token-level divergence for \emph{bounded credit
emphasis}: it adds local emphasis to the broadcast trajectory advantage without
claiming causal ground-truth credit or conserving advantage mass. Within each
response, our deterministic divergence/entropy segmenter uses
$\tilde d_{x,k,t}=(d_{x,k,t}-d_{\min})/(d_{\max}-d_{\min}+\epsilon_{\mathrm{norm}})$
with response extrema $d_{\min},d_{\max}$ and $\epsilon_{\mathrm{norm}}=0.1$.
A segment starts when $\tilde d_{x,k,t}>\lambda_{\mathrm{KL}}=0.15$ and extends
until entropy exceeds $\lambda_H=1.5$ times its onset value, subject to the
$0.2T$ cap in \suppref{sec:appendix}. Triggered-span tokens inherit the
onset saliency, while non-triggered tokens use their local $\tilde d_{x,k,t}$.
The saliency $s_{x,k,t}\in[0,1]$ marks confident, localized disagreement.
API/action spans are a safety/reporting slice, not the only segmentation rule.
The optimization mask includes only assistant-generated response tokens;
prompts, tool observations, simulator returns, and the credit-reference prefix
are excluded from the objective below.
Let $\mathcal{M}$ denote the resulting set of assistant-response token
positions.

\subsection{Credit-weighted policy gradient}
\label{sec:credit}
Let $A_{x,k}$ be the group-relative trajectory advantage and $\rho_{x,k,t}(\theta)$
the usual policy ratio. SGCD replaces the broadcast advantage with a
token-weighted advantage
\begin{equation}
  \begin{aligned}
  \widehat A_{x,k,t} &= A_{x,k} \, W_{x,k,t},\\
  s_{x,k,t} &= g(u_{x,k,t}),\\
  u_{x,k,t} &= \mathrm{sg}\!\left[d_{x,k,t},
  H(q_{x,k,t}^{K})\right],
  \end{aligned}
  \label{eq:weighted_adv}
\end{equation}
where $g$ is the deterministic divergence-spike and entropy-termination
segmenter from \S\ref{sec:segments}. In our implementation, the credit
multiplier is
\begin{equation}
  W_{x,k,t}
  = \mathrm{clip}\!\left(1+\gamma\,s_{x,k,t},
  1, c\right),
  \label{eq:weight_layer}
\end{equation}
where $\mathrm{clip}(a,\ell,u)=\min(\max(a,\ell),u)$, $c>1$, and
$\gamma\ge0$; reported runs use $\gamma{=}1$ and $c{=}2$. Low-saliency tokens
have $W\approx1$; high-saliency tokens amplify the existing advantage magnitude
without flipping its sign. The stop-gradient detaches feature construction from
actor and teacher scoring, and the multiplier is constant with respect to actor
parameters: there is no learned credit-weight network and no gradient through
$d_{x,k,t}$, $H(q^K)$, or a teacher-matching KL/JSD objective. Failed rollouts
receive larger negative advantage on high-saliency tokens; successful rollouts
receive concentrated positive credit; low-saliency tokens recover ordinary GRPO. The
actor loss is the standard
clipped surrogate with $\widehat A_{x,k,t}$. Let
$\bar\rho_{x,k,t}=\mathrm{clip}(\rho_{x,k,t},1-\epsilon,1+\epsilon)$; we use
symmetric $\epsilon{=}0.2$ (bounds $[0.8,1.2]$) as in PPO \citep{ppo2017}:
\begin{equation}
  \begin{aligned}
  \mathcal{L}_{\text{SGCD}}
  ={}& - \sum_{(x,k,t)\in\mathcal{M}} \\
     &\min\!\left(\rho_{x,k,t}\widehat A_{x,k,t},
      \bar\rho_{x,k,t}\widehat A_{x,k,t}\right).
  \end{aligned}
  \label{eq:sgcd_loss}
\end{equation}
This equation shows SGCD's credit-weighted PG component. Additional
regularization is recipe-specific and outside this notation. The distillation
pathway contributes no
teacher-matching actor-gradient term. The dense
teacher/student comparison and credit-reference construction determine
\emph{where} the verified advantage receives extra emphasis, not \emph{what}
token distribution the actor must imitate. The Vanilla GRPO comparators retain
their benchmark reference-KL regularizers: coefficient $0.05$ on
$\tau^3$-airline and $0.04$ on AppWorld, the latter following DeepSeekMath
\citep{deepseekmath2024}. SGCD omits reference-KL and instead uses detached
dense divergence as a replacement credit signal. This signal is not an
equivalent KL penalty because it has no direct teacher-matching actor-gradient
path. Finally, $\lambda_{\mathrm{KL}}=0.15$ in \S\ref{sec:segments} is a
normalized-divergence segment-onset threshold, not a loss coefficient.

\subsection{Why this removes the direct-SD failure path}
\label{sec:safety}
Earlier direct auxiliary variants have the form
$\mathcal{L}_{\text{PG}}+\lambda\mathcal{L}_{\text{SD}}$: if the teacher target
is wrong, privileged, or unreproducible, the auxiliary creates a second actor
gradient that can oppose the verifier. SGCD removes that path. The credit
reference and dense RKL are detached features; they identify where the verified
advantage receives bounded emphasis, while SGCD adds no distillation-derived
actor-gradient path beyond the reward-grounded PG term. This operationalizes the diagnosis in
\S\ref{sec:diagnosis}: use distillation for credit weighting, not to overrule
policy gradient.

\section{Experiments}
\label{sec:experiments}

\subsection{Setup}
We evaluate on AppWorld \citep{appworld2024} and $\tau^3$-airline, a corrected
airline task set in the $\tau$-bench family
\citep{taubench2024,tau2bench2025,tau3bench2026}. AppWorld reports the official
evaluator aggregates for Task Goal Completion (TGC) and Scenario Goal Completion
(SGC); $\tau^3$-airline reports
held-out mean Pass$^1$ with action-category diagnostics. AppWorld starts
from Qwen3.5-4B after diagnostic SFT regressed (\suppref{sec:sftdiag});
$\tau^3$-airline uses our SFT-warm-start ladder. Runtime settings and the
credit-reference prompt template are in \suppref{sec:appendix}. The AppWorld
Vanilla GRPO comparator uses matched rollout-generation hyperparameters and
evaluation settings and includes reference-KL; SGCD instead uses detached dense
divergence for credit weighting alongside its mixed-sibling construction.

\providecommand{\venuecompacttablefont}{\scriptsize}

\begin{table}[t]
  \centering
  \small
  \begin{tabular}{lcccc}
    \hline
    & \multicolumn{2}{c}{\textbf{test\_normal}} & \multicolumn{2}{c}{\textbf{test\_challenge}} \\
    \textbf{Method} & TGC & SGC & TGC & SGC \\
    \hline
    Base Qwen3.5-4B          & 23.8 & 10.7 & 11.3 & 2.2 \\
    Vanilla GRPO            & 42.9 & 16.1 & 24.7 & 7.2 \\
    \textbf{SGCD (ours)}     & \textbf{45.6} & \textbf{17.9} & \textbf{27.0} & \textbf{8.5} \\
    \hline
  \end{tabular}
  \caption{AppWorld held-out task-goal completion (TGC) and scenario-goal
    completion (SGC), reported as the official evaluator aggregates (\%), under
    greedy decoding with Qwen3.5-4B. \texttt{test\_normal}
    ($n{=}168$ tasks / $56$ scenarios) and \texttt{test\_challenge}
    ($n{=}417$ / $139$) are reported separately. SGCD raises official TGC over
    the un-finetuned base and gives higher held-out point
    estimates than the Vanilla GRPO control under matched rollout-generation
    hyperparameters, environment, and evaluation settings. This recipe-level
    comparison uses reference-KL only
    in Vanilla GRPO; SGCD instead uses detached divergence for credit weighting.
    AppWorld values are single-training-run point
    estimates; time and compute constraints prevented multi-seed reruns.}
  \label{tab:appworld}
\end{table}

\begin{table}[t]
  \centering
  \venuecompacttablefont
  \setlength{\tabcolsep}{2.2pt}
  \begin{tabular}{lcccc}
    \hline
    \textbf{Method} & \textbf{step} & \textbf{mean Pass$^1$} & \textbf{action succ.} & \textbf{info succ.} \\
    \hline
    SFT base              & 800 & 0.438 & 26.9 & 68.8 \\
    SDPO                  & 300 & 0.317 & 0.0 & 91.7 \\
    Vanilla GRPO         & 120 & 0.583 & 41.0 & 91.7 \\
    \textbf{SGCD (ours)}  & 120 & \textbf{0.617} & \textbf{44.9} & \textbf{93.8} \\
    \hline
  \end{tabular}
  \caption{$\tau^3$-airline held-out mean Pass$^1$ and per-category
    diagnostics over 20 task specifications. Each task uses a paired grid of
    three evaluation seeds and four greedy trials per seed. Pass$^1$ is the
    macro-average of per-task binary success rates; because every task has 12
    episodes, it also equals total successes divided by 240. \emph{Action succ.} is the state-changing-action
    diagnostic; \emph{info succ.} is the information-only diagnostic. The SGCD
    row is the closest-to-mean run ($148/240$); Table~\ref{tab:seedrobust}
    reports the three-training-run summary, and the tau margin is directional rather
    than variance-controlled. Vanilla GRPO uses the benchmark reference-KL
    regularizer with coefficient $0.05$; SGCD omits reference-KL and uses
    detached divergence for credit weighting.}
  \label{tab:tau3}
\end{table}

\begin{table}[t]
\centering
\small
\setlength{\tabcolsep}{4pt}
  \begin{tabular}{llcc}
\toprule
  Metric & Anchor & SGCD (mean$\pm$std) & min $\Delta$ \\
\midrule
$\tau^3$ mean Pass$^1$  & 0.583 & $0.615\pm0.006$ & $+0.025$ \\
  \bottomrule
\end{tabular}
\caption{$\tau^3$-airline SGCD seed variation against the fixed single-run
Vanilla GRPO anchor. SGCD uses three independent training runs; its standard
deviation is SGCD-only dispersion, not between-method variance. \emph{min
$\Delta$} is the worst-seed margin over the anchor. We make no significance
claim; action/information diagnostics appear only in Table~\ref{tab:tau3}.}
\label{tab:seedrobust}
\end{table}

\subsection{Main results}
On AppWorld (Table~\ref{tab:appworld}), SGCD raises official TGC over the
un-finetuned base on both splits (\texttt{test\_normal} TGC $23.8\to45.6$;
\texttt{test\_challenge} TGC $11.3\to27.0$) and gives higher point estimates than
the Vanilla GRPO control under matched rollout-generation hyperparameters,
environment, and evaluation settings ($+2.7$ and $+2.3$ TGC). This is a
recipe-level comparison rather than
an isolated component test because only Vanilla GRPO includes reference-KL.
The margin is modest; SGCD adds bounded credit weighting
inside a reward-grounded policy-gradient recipe.

On $\tau^3$-airline (Table~\ref{tab:tau3}), the benchmark where we first
diagnosed the failure, three SGCD training runs give a higher held-out mean Pass$^1$
summary ($0.615\pm0.006$) than the Vanilla GRPO comparator ($0.583$).
The detailed diagnostic row reports the closest-to-mean SGCD run
($148/240=0.617$): its state-changing-action diagnostic remains nonzero ($44.9\%$
vs.\ $41.0\%$ for Vanilla GRPO), while the historical SDPO diagnostic run has
collapsed to $0\%$. We read the SGCD--Vanilla comparison as directional evidence
under their matched setup, not as a variance-controlled effect; the comparator
is a fixed single-run anchor and we
make no between-method significance claim (Table~\ref{tab:seedrobust}).

\subsection{Ablations}
\label{sec:ablations}
Table~\ref{tab:ablation} removes one component at a time. Removing the sibling
credit reference lowers \texttt{test\_normal} TGC from $45.6$ to $44.0$.
Replacing divergence weights with mean-matched uniform weights lowers it to
$43.8$, so the effect is not explained only by average multiplier size. The
direct/uncapped SD auxiliary uses the same teacher/student divergence as an
imitation loss and drops to $38.2$, below Vanilla GRPO at $42.9$; we treat this
as an untuned diagnostic, not a tuned baseline. Vanilla GRPO contains no SGCD
component.

\begin{table}[t]
  \centering
  \venuecompacttablefont
  \setlength{\tabcolsep}{2.5pt}
  \begin{tabular}{@{}lcc@{}}
    \toprule
    \textbf{Configuration} & \textbf{TGC norm.} & \textbf{TGC chal.} \\
    \midrule
    \textbf{SGCD (full)}                         & \textbf{45.6} & \textbf{27.0} \\
    \quad no sibling credit reference             & 44.0 & 25.4 \\
    \quad mean-matched uniform weights            & 43.8 & 25.3 \\
    \quad direct/uncapped SD aux.                  & 38.2 & 20.6 \\
    \midrule
    Vanilla GRPO                                   & 42.9 & 24.7 \\
    \bottomrule
  \end{tabular}
  \caption{Ablating SGCD on AppWorld held-out TGC (\%). Removing the sibling
    credit reference or replacing divergence weights with mean-matched uniform
    weights gives lower point estimates;
    replacing credit reweighting with a diagnostic direct SD auxiliary gives
    lower TGC than the reported Vanilla GRPO control.}
  \label{tab:ablation}
\end{table}

\subsection{Analysis}
During SGCD training, dense reverse-KL, teacher entropy, selected segments,
weight distributions, and clipping are monitored as sanity checks. They are not
headline evidence, and the inference-time student carries no external
dependency.
\FloatBarrier

\section{Conclusion}
\label{sec:conclusion}

In the recipes studied here, naive self-distillation can collapse multi-turn
tool use: on $\tau^3$-airline,
easy information-only behavior remains while state-changing action success falls
to zero. SGCD instead uses distillation only to construct bounded credit weights
for the policy-gradient term and adds no teacher-matching actor gradient. AppWorld supplies the main
positive held-out result; $\tau^3$-airline supplies a directional held-out
comparison and the diagnosis of the direct-SD failure mode. The evidence
supports a narrow design recommendation: use distillation to shape
verifier-grounded credit while avoiding a competing teacher-matching imitation
path in these settings.

\section*{Limitations}
\label{sec:limitations}

\paragraph{Single student-model family and scale.} All trained student policies
use Qwen3.5-4B. Generalization across student-model families and to larger scales
(e.g.\ 8B) is future work.

\paragraph{Benchmarks.} SGCD is evaluated on two benchmarks (AppWorld and
$\tau^3$-airline). Broader coverage across additional agentic domains is future
work.

\paragraph{External-model dependence.} The credit reference is
produced by \creditreferenceproviderphrase{} during training. Its cost,
availability, and quality are practical constraints. The inference-time student has no
such dependency, but training-time quality still matters because poor credit
references can produce weak or noisy credit features. We do not claim that the
reference model choice is immaterial. The credit-reference model family and the
structured prompt used to elicit it are described in \suppref{sec:appendix}. The
$\tau^3$ data generation, training environment, and held-out evaluation also use
an external instruction-following user simulator shared across compared arms;
this is an evaluation dependency, not a component of the deployed student.

\paragraph{Credit reference is not formal ground truth.} SGCD approximates
stepwise reference behavior from verified siblings. This is weaker than a
benchmark-provided stepwise oracle and can miss valid alternative plans. The
bounded credit weights reduce the risk of overcommitting to one interpretation,
but they do not make the credit reference infallible.

\paragraph{Training-run coverage.} AppWorld values are single-run point estimates because time and compute constraints prevented multi-seed reruns, so we make no AppWorld variance or significance claim. On $\tau^3$-airline, SGCD uses three independent training runs (Table~\ref{tab:seedrobust}), but Vanilla GRPO is a fixed single-run anchor; the reported dispersion is therefore SGCD-only and the comparison remains directional rather than variance-controlled.

\paragraph{AppWorld recipe comparison.} The AppWorld arms match rollout,
environment, and evaluation settings, but not their regularization term:
only Vanilla GRPO includes reference-KL. SGCD instead uses detached dense
divergence for credit weighting, so the result compares complete recipes and
does not isolate the effect of removing reference-KL.

\paragraph{Diagnostic and ablation scope.} The AppWorld SFT regression table is
a diagnostic subset result, not a headline held-out result. The full SGCD result
also depends on the exact external-LLM prompt and dynamic-sampling rollout set;
we report the core controls but do not exhaustively sweep every prompt-builder
or divergence variant.

\paragraph{Artifact availability.} The review package does not include the full
training/evaluation code, generated training trajectories and corrected task
artifacts, or exact run-level manifests. The supplement reports the available
settings, prompt template, and evaluation definitions, but the submitted
materials do not support exact end-to-end replication.

\section*{Ethical Considerations}

SGCD is a training-time method for tool-use agents and introduces no new data
collection involving human subjects. The experiments build on public benchmark
resources. The method's
privileged signal, summaries of on-policy sibling rollouts from
\creditreferenceproviderphrase{}, is
confined to training; the inference-time model receives no credit reference, sibling
evidence, oracle, or external model. The sanitizer excludes credentials,
evaluator internals, hidden gold actions, and answer-bearing
deployment-invisible content from the external summarizer. As with any agent that executes actions in real
applications, deployment should retain human oversight for irreversible
operations. SGCD only changes training-time credit weighting; it is not a
substitute for application-level safety controls.

Generative AI systems assisted with language editing, consistency checks,
citation verification, and preparation of the conceptual method schematic.
They were not treated as authors or as sources of scientific evidence. The
authors independently reviewed the final text, references, figures, and claims
and remain responsible for the manuscript in full.

\bibliography{refs}

\providecommand{\venuecompacttablefont}{\scriptsize}
\appendix

\section{SGCD Objective: Bounded Credit Reweighting}
\label{sec:theory}

We record the properties used in \S\ref{sec:method}. SGCD does not optimize a
standalone KL/JSD actor loss. Instead, it uses detached dense divergence
features to produce a bounded token-level multiplier on the policy-gradient
advantage. Let $j=(i,t)$ index a masked response token and let $A_i$ be the
trajectory-level group-relative advantage. Let
\[
  \begin{aligned}
  r_j(\theta)
  &=
  \frac{\pi_\theta(y_j\mid x_i,y_{i,<t})}
       {\pi_{\mathrm{old}}(y_j\mid x_i,y_{i,<t})},\\
  \bar r_j(\theta)
  &=
  \mathrm{clip}(r_j(\theta),1-\epsilon,1+\epsilon).
  \end{aligned}
\]
Reported runs use $\epsilon{=}0.2$ (bounds $[0.8,1.2]$).
The ordinary clipped GRPO token loss is
\[
  \ell_j^{\mathrm{GRPO}}(\theta)
  =
  -\min\{r_j(\theta)A_i,\bar r_j(\theta)A_i\}.
\]
SGCD computes detached credit features $u_j=\mathrm{sg}[d_j,H_j]$, where
$d_j$ is the dense teacher/student divergence and $H_j$ is teacher entropy.
The deterministic segmenter gives $s_j=g(u_j)$, and the bounded multiplier
satisfies $1\le W_j\le c$. In the reported implementation,
\[
  W_j=\mathrm{clip}(1+\gamma s_j,1,c),
\]
with $\gamma{=}1$ and $c{=}2$. The SGCD token loss is
\[
  \ell_j^{\mathrm{SGCD}}(\theta)
  =
  -\min\{r_j(\theta)A_iW_j,\bar r_j(\theta)A_iW_j\}.
\]
The propositions below concern this component; recipe-specific regularizers are
outside the notation.

\paragraph{Proposition 1 (no teacher-gradient path).}
Since $u_j$ is stop-gradient, $\partial W_j/\partial\theta=0$ for actor
parameters $\theta$. Therefore the SGCD credit-weighting path adds gradients
only through the reweighted policy ratio $r_j(\theta)$; it adds no
distillation-derived actor-gradient term of the form
\[
  \nabla_\theta D(\pi_\theta\Vert q_{\mathrm{teacher}}).
\]
Dense RKL and entropy are credit features, not a direct actor-gradient
distillation loss.

\paragraph{Proposition 2 (positive scaling of GRPO).}
For every token $j$,
\[
  \ell_j^{\mathrm{SGCD}}(\theta)
  =
  W_j\,\ell_j^{\mathrm{GRPO}}(\theta).
\]
Indeed, because $W_j>0$,
\[
  \min\{r_jA_iW_j,\bar r_jA_iW_j\}
  =
  W_j\min\{r_jA_i,\bar r_jA_i\}.
\]
Since $W_j$ is detached from actor parameters,
\[
  \nabla_\theta \ell_j^{\mathrm{SGCD}}(\theta)
  =
  W_j\nabla_\theta \ell_j^{\mathrm{GRPO}}(\theta).
\]
These equalities are pointwise where the clipped surrogate is differentiable;
at clip or minimum tie points, the same positive scaling holds for any
consistent subgradient choice because $W_j$ is detached and positive.

\paragraph{Corollary 1 (sign preservation).}
The reshaped advantage $\widehat A_j=A_iW_j$ satisfies
\[
  \mathrm{sign}(\widehat A_j)=\mathrm{sign}(A_i),
  \qquad
  |\widehat A_j|\le c|A_i|.
\]
Thus SGCD can scale local credit but cannot flip the sign of the
verifier-grounded advantage.

\paragraph{Corollary 2 (tokenwise PG alignment).}
For every token $j$,
\[
  \left\langle
  \nabla_\theta \ell_j^{\mathrm{SGCD}},
  \nabla_\theta \ell_j^{\mathrm{GRPO}}
  \right\rangle
  =
  W_j
  \left\|
  \nabla_\theta \ell_j^{\mathrm{GRPO}}
  \right\|^2
  \ge 0.
\]
Hence SGCD is tokenwise aligned with the corresponding GRPO update: it either
scales the same clipped-surrogate policy-gradient direction or, in a flat
clipped region, scales a zero gradient.

\paragraph{Corollary 3 (bounded contribution envelope).}
For every token,
\[
  \begin{aligned}
  |\ell_j^{\mathrm{SGCD}}|
  &\le c|\ell_j^{\mathrm{GRPO}}|,\\
  \|\nabla_\theta \ell_j^{\mathrm{SGCD}}\|
  &\le c\|\nabla_\theta \ell_j^{\mathrm{GRPO}}\|.
  \end{aligned}
\]
For a batch $B$,
\[
  \|\nabla_\theta L_{\mathrm{SGCD}}\|
  \le
  c\sum_{j\in B}
  \|\nabla_\theta \ell_j^{\mathrm{GRPO}}\|.
\]
This is a bound relative to the tokenwise policy-gradient contribution
envelope. It does not require, and does not imply, that SGCD's batch gradient is
within a factor $c$ of the realized GRPO batch gradient, because ordinary token
gradients may cancel and the bounded reweighting can change that cancellation.

\paragraph{Corollary 4 (bounded local movement under smooth KL).}
Assume the average policy KL is locally smooth, so that for sufficiently small
updates $\Delta\theta$,
\[
  \mathrm{KL}(\pi_\theta\Vert\pi_{\theta+\Delta\theta})
  \le
  \frac{L_{\mathrm{KL}}}{2}\|\Delta\theta\|^2 .
\]
For one SGCD step $\Delta\theta=-\eta\nabla_\theta L_{\mathrm{SGCD}}$,
\[
  \begin{aligned}
  &\mathrm{KL}(\pi_\theta\Vert\pi_{\theta+\Delta\theta})\\
  &\quad\le
  \frac{L_{\mathrm{KL}}}{2}
  \eta^2c^2
  \left(
  \sum_{j\in B}
  \|\nabla_\theta \ell_j^{\mathrm{GRPO}}\|
  \right)^2 .
  \end{aligned}
\]
This is not a global convergence or trust-region theorem. It states that SGCD's
policy movement is bounded by a clipped, detached reweighting of
verifier-grounded policy-gradient token contributions, rather than by an
independent teacher-matching actor loss.

\section{Reproducibility Details}
\label{sec:appendix}

\paragraph{Shared runtime configuration.} The AppWorld arms share the runtime
in Table~\ref{tab:recipe}, but only Vanilla GRPO uses reference-KL, with
coefficient $0.04$. The matched $\tau^3$-airline Vanilla GRPO and SGCD arms
share Table~\ref{tab:recipe_tau3}, including the same rollout, environment, and
evaluation harness, but only Vanilla GRPO uses reference-KL, with coefficient
$0.05$. The earlier SDPO collapse
run used a different diagnostic training recipe and is not covered by that
runtime table; its reported checkpoint was re-evaluated with the same canonical
held-out harness. The base student model is Qwen3.5-4B. These tables list shared
environment and sampling settings; method-specific hyperparameters are given in
\S\ref{sec:method} and \S\ref{sec:theory}.

\begin{table}[h]
  \centering
  \venuecompacttablefont
  \setlength{\tabcolsep}{3pt}
  \begin{tabular}{p{0.46\columnwidth}p{0.42\columnwidth}}
    \hline
    \textbf{Runtime setting} & \textbf{Value} \\
    \hline
    Base model                  & Qwen3.5-4B (raw) \\
    Rollouts per prompt ($n$)   & $8$ \\
    Train / mini-batch          & $8$ / $8$ \\
    PPO epochs                  & $1$ \\
    Learning rate               & $1\mathrm{e}{-6}$ \\
    Max prompt / response       & $16{,}384$ / $16{,}384$ tokens \\
    Max model length            & $32{,}768$ tokens \\
    Max user / assistant turns  & $30$ / $30$ \\
    Max observation             & $12{,}000$ chars \\
    Env steps per task          & $30$ \\
    Thinking enabled            & yes \\
    Rollout temp. / top-$p$ (train) & $1.0$ / $1.0$ \\
    Rollout temp. / top-$p$ (val)   & $1.0$ / $0.95$ \\
    Decoding (eval)             & greedy \\
    GPUs / tensor parallel      & $8$ / $1$ \\
    RL train/dev subset; held-out eval & $90$ / $57$ tasks; \texttt{test\_normal} $168$ + \texttt{test\_challenge} $417$ \\
    \hline
  \end{tabular}
  \caption{Shared runtime and environment configuration for the AppWorld arms.
    SGCD and the comparators use matched rollout-generation hyperparameters,
    evaluation settings, and the same environment/evaluator; only Vanilla GRPO
    includes reference-KL. SGCD
    instead uses detached divergence and mixed-sibling credit construction;
    Vanilla GRPO does not use mixed-group filtering. The response budget of $16{,}384$ tokens
    is shared across up to $30$ assistant turns; observations are truncated at
    $12{,}000$ characters. The train/dev counts are the RL subset used in these
    arms, not benchmark-wide official split counts. Held-out evaluation uses
    \texttt{test\_normal} ($168$ tasks / $56$ scenarios) and
    \texttt{test\_challenge} ($417$ / $139$) separately.}
  \label{tab:recipe}
\end{table}

\paragraph{Dynamic-sampling exhaustion.} The reported AppWorld and
$\tau^3$-airline SGCD runs use the same rule. Each candidate prompt group
contains eight siblings. We retain only complete mixed groups, targeting eight
kept groups per optimizer update, and refill for at most ten generation batches.
All-success and all-failure groups are discarded. If the target is still unmet
at the cap, the optimizer update is skipped and training continues.

\paragraph{$\tau^3$-airline runtime.} The matched Vanilla GRPO and SGCD arms
share the runtime in Table~\ref{tab:recipe_tau3}. Unlike the AppWorld arms, the $\tau^3$ policy is
initialized from a supervised checkpoint, which the diagnosis tracks as the
PG-dominant control (\S\ref{sec:diagnosis}). The interaction is bounded to $16$
user/assistant turns, and each tool/user observation is truncated to $512$
characters.

\begin{table}[h]
  \centering
  \venuecompacttablefont
  \setlength{\tabcolsep}{3pt}
  \begin{tabular}{p{0.46\columnwidth}p{0.42\columnwidth}}
    \hline
    \textbf{Runtime setting} & \textbf{Value} \\
    \hline
    Base model                  & Qwen3.5-4B, SFT-initialized \\
    Rollouts per prompt ($n$)   & $8$ \\
    Train / mini-batch          & $8$ / $8$ \\
    PPO epochs                  & $1$ \\
    Learning rate               & $1\mathrm{e}{-6}$ \\
    Max prompt / response       & $16{,}384$ / $16{,}384$ tokens \\
    Max model length            & $32{,}768$ tokens \\
    Max user / assistant turns  & $16$ / $16$ \\
    Max chars / observation     & $512$ \\
    Thinking enabled            & yes \\
    Rollout temp. / top-$p$ (train) & $0.4$ / $0.95$ \\
    Rollout temp. / top-$p$ (val)   & $0.4$ / $0.95$ \\
    Decoding (eval)             & greedy; $3$ seeds $\times$ $4$ trials \\
    GPUs / tensor parallel      & $8$ / $1$ \\
    Domain / held-out split     & airline / 20 task specs \\
    \hline
  \end{tabular}
  \caption{Shared runtime and environment configuration for the matched
    $\tau^3$-airline Vanilla GRPO and SGCD arms. The policy is SFT-initialized;
    Vanilla GRPO uses frozen-SFT low-variance reference-KL with fixed coefficient
    $0.05$, whereas SGCD omits reference-KL and uses detached divergence for
    credit weighting. The SDPO diagnostic training recipe is excluded from this
    table.}
  \label{tab:recipe_tau3}
\end{table}

\paragraph{$\tau^3$-airline user simulator.} All $\tau^3$-airline experiments
use the same instruction-following user simulator across arms. It role-plays the
airline customer according to the task specification. The same user-simulator
model and configuration are shared by the
SFT data generation, the Vanilla GRPO anchor, the SDPO collapse runs, and SGCD, so
the comparison introduces no method-specific user-simulator configuration.

\paragraph{$\tau^3$-airline SFT data and training.} After task-success and
format/trace validation, we retained $9{,}195$ full multi-turn trajectories
generated deterministically by \sftteacherproviderphrase{} paired with the same
instruction-following user simulator. A task-level split assigned $7{,}927$
trajectories from 25 tasks to SFT training and $1{,}268$ trajectories from three
tasks to SFT validation; each trajectory was one unpacked example. We trained
Qwen3.5-4B with full-parameter cross-entropy, maximum sequence length $32{,}768$,
global batch size $8$ (micro-batch $1$ per GPU), and constant learning rate
$1\mathrm{e}{-5}$ without warmup. The run was configured for two epochs
($990$ optimizer updates per epoch; $1{,}980$ planned) and stopped after saving
checkpoint step $1{,}600$ (1.62 epochs). All RL arms use the intermediate
step-$800$ checkpoint (0.81 epoch), selected from the step-$600$--$800$
validation-loss trough and subsequent rollout checks. It reaches held-out mean
Pass$^1$ $0.438$ on the 20-task-specification split
(Table~\ref{tab:tau3}), establishing the tool-use floor that naive SDPO
subsequently collapses and that PG-dominant training preserves.

\paragraph{Evaluation protocol.} AppWorld is scored with the benchmark's
state-based evaluator, reporting TGC and SGC \emph{separately} for
\texttt{test\_normal} ($168$ tasks / $56$ scenarios) and
\texttt{test\_challenge} ($417$ / $139$) under greedy decoding. These headline
values are the official evaluator aggregates, not success fractions reconstructed
from integer counts. The splits are not combined; the earlier
\texttt{batch\_001} SFT diagnostic in
\S\ref{sec:sftdiag} is a separately labeled pooled subset. The
$\tau^3$-airline diagnosis uses 20 held-out
task specifications (13 state-changing action tasks, 4 information-only tasks,
1 transfer, 2 no-action tasks). Evaluation uses a paired grid of three evaluation seeds
and four greedy trials per seed, giving 12 replay episodes per task specification
and 240 replay episodes total. Table~\ref{tab:tau3} reports binary success
(final evaluator reward $\geq1.0$), macro-averaged across task specifications.
Equal 12-episode task counts make this identical to total successes divided by
240. The representative SGCD row is 148/240 overall, 70/156 on
state-changing-action tasks, and 45/48 on information-only tasks.
In our internal ledgers, \texttt{write\_req} denotes state-changing action
tasks and \texttt{read\_only} denotes information-only tasks.
The same user simulator (strong instruction-following) drives the customer side
for both training rollouts and held-out evaluation. The zero-tool-collapse audit
measures, per training step, the fraction of information-only successes that
invoke no tool and the mean number of tools per information-only success. This
historical success-conditioned audit is distinct from
Figure~\ref{fig:tau_training_diag}, whose top panel reports mean tool count over
all diagnostic rollouts.

\paragraph{pass$^k$ vs.\ pass@$k$ (formal definition).}
We follow the $\tau$-bench evaluation protocol \citep{taubench2024}. The two
metrics have \emph{opposite} monotonicity in $k$:
\begin{itemize}
  \item \textbf{pass@$k$} (standard code-generation convention): draw $k$
    independent rollouts; a task is solved if \emph{at least one} succeeds:
    \[
      \mathrm{pass@}k = \mathbb{E}\!\left[
        1 - \prod_{i=1}^{k}(1-s_i)
      \right],
    \]
    where $s_i\in\{0,1\}$ is the success indicator of rollout $i$. Higher $k$
    \emph{increases} pass@$k$ (more attempts $\Rightarrow$ higher chance of
    at least one success).
  \item \textbf{pass$^k$} ($\tau$-bench convention): draw $k$ independent
    rollouts; a task is solved only if \emph{all $k$} succeed:
    \[
      \mathrm{pass}^k = \mathbb{E}\!\left[
        \prod_{i=1}^{k} s_i
      \right].
    \]
    Higher $k$ \emph{decreases} pass$^k$ (more attempts $\Rightarrow$ stricter
    consistency requirement). pass$^k$ measures \emph{reliability}: can the
    agent solve the task \emph{every time}, not just once?
\end{itemize}
In Table~\ref{tab:tau3} and throughout this paper, held-out \textbf{mean
Pass$^1$} is the task macro-average defined above; equal 12-episode task counts
make it identical to the 240-episode binary micro-average. The $\tau$-bench authors
motivate pass$^k$ as a
reliability metric for deployed agents where consistency matters more than
best-of-$k$ luck \citep{taubench2024}.

\paragraph{Credit-reference construction.} For a mixed sibling group, the
external LLM receives the policy-visible task instruction, rollout-visible
sibling traces, and each sibling's binary verifier outcome after sanitization. If
the entire set fits within context, all siblings are shown. If it is too long,
we include the smaller outcome class and match it with the opposite class. When
there are more successes than failures, we prefer shorter successful siblings;
when there are more failures than successes, we prefer longer failed siblings.
This keeps concise solutions and informative failures while staying within the
external LLM context. The reference is rebuilt from the current on-policy
siblings each time the task is sampled; we do not use a persistent skill bank.
The credit reference is generated by \creditreferenceproviderphrase{}, prompted with the
structured credit-scoring template in Table~\ref{tab:credit_prompt}.
Table~\ref{tab:external_models} summarizes the external components used only at
training time.

\begin{table}[h]
  \centering
  \venuecompacttablefont
  \setlength{\tabcolsep}{3pt}
  \begin{tabular}{p{0.28\columnwidth}p{0.62\columnwidth}}
    \hline
    \textbf{Component} & \textbf{Operational description} \\
    \hline
    Credit-reference LLM &
      \creditreferenceproviderphrase{} used only during training. Input is the
      sanitized task instruction plus mixed sibling traces and verifier outcomes;
      output follows Table~\ref{tab:credit_prompt}. Decoding, timeout, retry,
      and call/token accounting are implementation metadata. \\
    SFT trajectory teacher &
      \sftteacherproviderphrase{} used before RL to generate retained
      successful supervised-initialization trajectories with the same user
      simulator. Failed trajectories are discarded and the teacher is absent
      from RL rollouts and deployment. \\
    $\tau^3$ user simulator &
      \simulatorproviderphrase{} and task-specification prompt template
      across SFT data generation, Vanilla GRPO diagnostics, SDPO diagnostics, SGCD training, and
      held-out evaluation. No method-specific simulator prompt changes. \\
    \hline
  \end{tabular}
  \caption{Training-time external components. They are absent from deployment:
    the final student receives only the clean task prompt.}
  \label{tab:external_models}
\end{table}

\begin{table*}[t]
  \centering
  \venuecompacttablefont
  \setlength{\tabcolsep}{4pt}
  \begin{tabular}{p{0.20\textwidth}p{0.72\textwidth}}
    \hline
    \textbf{Prompt block} & \textbf{Template content} \\
    \hline
    Role &
    You are a training-time credit analyst. Given one task and several
    policy-generated sibling rollouts with verifier outcomes, write a concise
    stepwise credit reference. \\
    Input &
    Task instruction; visible tool/API traces and observations for each
    sibling; each sibling's success/failure outcome; rollout-visible alignment
    metadata such as turn/order identifiers. \\
    Rules &
    Use only information present in the visible traces. Do not quote hidden
    labels, evaluator reports, credentials, or final answer literals. Prefer
    abstract state checks and action choices over copying task-specific values. \\
    Output &
    (1) state checks that successful siblings establish; (2) reusable
    successful action branches; (3) failed branches and why they appear wrong;
    (4) likely deviation points where failures depart from successful evidence;
    (5) literals or values that should remain masked during credit scoring. \\
    Use &
    Insert this reference only in the teacher-side scoring prefix during
    training. Never insert it into the student rollout prompt or deployment
    prompt. \\
    \hline
  \end{tabular}
  \caption{Credit-reference prompt template. The template is intentionally
    benchmark-generic: sibling traces provide the evidence, while the prompt
    asks the external LLM to abstract reusable state/action credit rather than
    copy hidden answers.}
  \label{tab:credit_prompt}
\end{table*}

\paragraph{Leakage and literal masking.} Before calling the external LLM, we
remove benchmark-private supervision and values that should not be recoverable
from the clean deployment prompt. Concretely, the sanitizer masks evaluator or
oracle reports, hidden gold actions, credentials, payment or account values,
and answer-bearing literals. This rule is not AppWorld-specific: for any
tool-use benchmark, the credit reference may summarize policy-visible behavior
and observations, but it may not expose private labels or task answers.

\paragraph{Dense credit features.} We project both the clean-student distribution
and the credit-reference-conditioned teacher distribution onto the same local
support containing the teacher top-$100$ tokens, the observed token, and a tail bucket.
The reported results use reverse-KL; no alternative-divergence ablation is
claimed. Divergence spikes and teacher
entropy define credit saliency before the bounded credit multiplier produces
token weights.

\paragraph{Credit segmenter.} SGCD uses a deterministic reverse-KL/entropy
segmenter with sibling-derived credit-reference teacher context rather than a
ground-truth-conditioned teacher. For each assistant response, let $d_{\min}$
and $d_{\max}$ denote the minimum and maximum dense reverse-KL. We normalize
$d_t$ as $\tilde d_t=(d_t-d_{\min})/(d_{\max}-d_{\min}+\epsilon_{\mathrm{norm}})$,
using $\epsilon_{\mathrm{norm}}=0.1$. A segment starts when
$\tilde d_t>\lambda_{\mathrm{KL}}=0.15$ and extends until either entropy exceeds
$\lambda_H=1.5$ times its onset value or its length reaches $0.2T$, where $T$ is
the rollout length. There is no additional minimum-length constraint. Tokens inside a triggered segment receive the
onset saliency; tokens outside triggered segments use their local
$\tilde d_t$. There is no learned saliency network. The final multiplier remains
bounded by Eq.~\ref{eq:weight_layer}, with $\gamma{=}1$ and $c{=}2$.

\paragraph{Response and credit masks.} Policy-gradient loss is computed only on
assistant-generated response tokens. System prompts, user/task prompts, API
documentation, tool observations, simulator returns, and the credit-reference
prefix itself are masked out. Generated reasoning tokens remain model output and
can receive the ordinary PG loss; SGCD may also use their detached divergence as
credit evidence. API/action spans are additionally logged as a safety and
analysis slice, but fixed tool-call boundaries are not the only credit segments.
Answer-bearing literals and masked values are excluded from credit-reference
scoring; for response tokens that remain in the policy-gradient mask, SGCD
saliency is zero and therefore $W=1$.

\paragraph{Runtime monitoring.} Dense teacher scoring adds an extra forward
path, so we monitor step time, dense-scoring time, selected dense-token counts,
credit-token counts, and peak GPU memory. The run is considered invalid if the
dense features are all zero, the credit weights are inert, response masks include
prompt/tool-observation tokens, or the detached divergence path becomes a direct
actor-gradient KL/JSD loss.

\section{Additional Result Tables}
\label{sec:appendix_tables}

The main body reports headline results (Tables~\ref{tab:appworld}--\ref{tab:tau3}),
the seed summary (Table~\ref{tab:seedrobust}), and the ablation grid
(Table~\ref{tab:ablation}). This section collects the
self-distillation method family comparison (Table~\ref{tab:sdfamily}) for
reference: it positions SGCD's use of distillation as detached credit features
rather than a standalone actor-gradient loss, the structural distinction from
prior SD methods.

\paragraph{Direct-SD diagnostic row.} The direct/uncapped SD row uses the same
credit-reference teacher context, top-$100$ local support, and assistant-response
mask as SGCD, but feeds the teacher/student divergence into a standalone
teacher-matching auxiliary of the form
$\mathcal{L}_{\mathrm{PG}}+\lambda\mathcal{L}_{\mathrm{SD}}$
rather than the bounded detached multiplier. It is the controlled, in-codebase
analogue of the SDPO-style direct self-distillation failure in \S\ref{sec:diagnosis}.
Relative to SGCD's ablation setting, the salient change is that the divergence
enters as an actor gradient instead of a detached credit weight. We report it as
a single untuned diagnostic setting with a lower point estimate than the reported
Vanilla GRPO, not as a swept baseline; we did not tune $\lambda$ or its
schedule. The archived run/evaluation ledger does not retain the numerical
coefficient or schedule, so this row should be interpreted only as a diagnostic
comparison and not as a fully reproducible baseline.

\begin{table*}[t]
  \centering
  \venuecompacttablefont
  \setlength{\tabcolsep}{4pt}
  \begin{tabular}{p{0.13\textwidth}p{0.23\textwidth}p{0.31\textwidth}p{0.21\textwidth}}
    \hline
    \textbf{Method} & \textbf{Objective role} & \textbf{Teacher / privileged context} & \textbf{Signal used} \\
    \hline
    SDPO \citep{sdpo2026}      & feedback-conditioned SD         & rich feedback / successful samples & sym.\ JSD or RKL; student-top-$K$ + tail \\
    OPSD \citep{opsd2026}      & standalone reward-free SD       & label-privileged trace / answer & full-vocab forward KL; JSD-$\beta$ ablated \\
    Skill-SD \citep{skillsd2026} & GRPO $+\lambda$ SD loss       & trajectory-derived NL skill summaries & imp.-weighted reverse-KL \\
    SDAR \citep{sdar2026}      & GRPO $+\lambda$ gated SD aux    & retrieved skill / privileged context & $\sigma(\beta\Delta)$ gap gate \\
    \textbf{SGCD (ours)}       & \textbf{PG credit reweighting}  & \textbf{mixed siblings summarized into a credit reference} & \textbf{dense RKL+entropy, detached credit weights} \\
    \hline
  \end{tabular}
  \caption{Self-distillation method family. SGCD uses distillation differently:
    teacher/student divergence is a detached credit-assignment signal for the
    policy-gradient surrogate, not a standalone actor-gradient auxiliary.}
  \label{tab:sdfamily}
\end{table*}

\section{Training-Trajectory Diagnostics}
\label{sec:training_trajectory_diagnostics}

\begin{figure*}[t]
  \centering
  \includegraphics[width=0.96\textwidth]{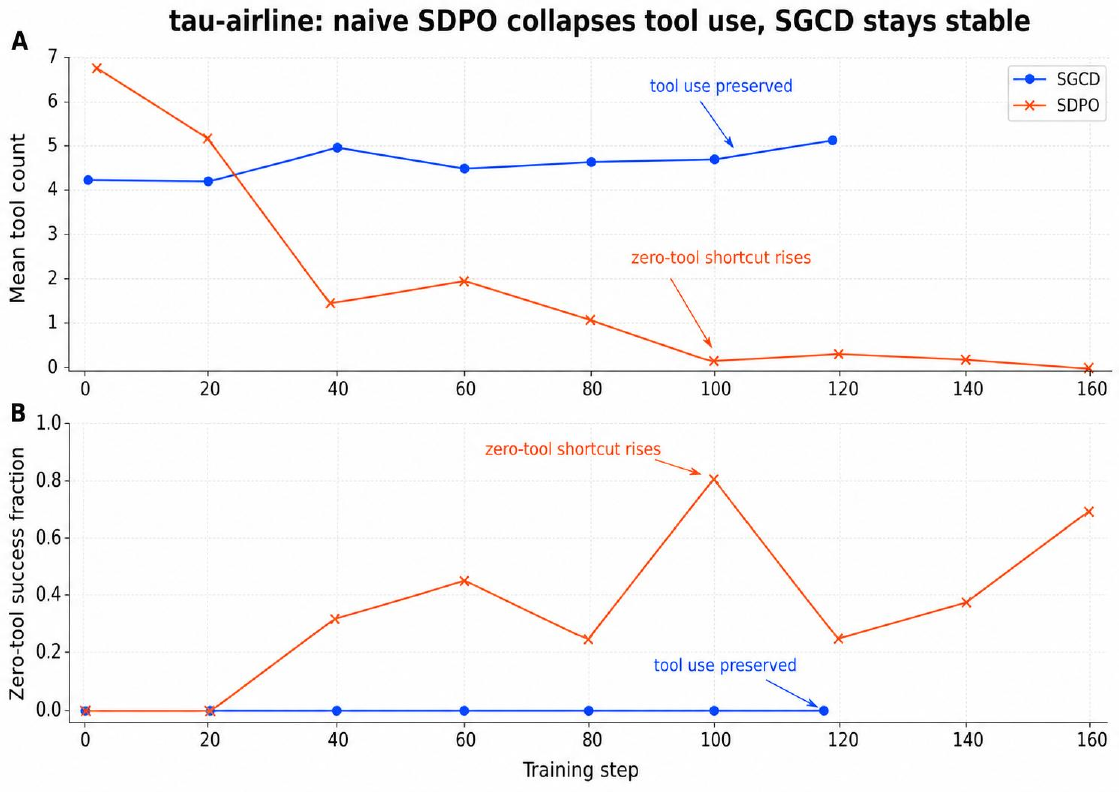}
  \caption{$\tau^3$-airline training diagnostic trajectories. In this diagnostic
    trace, SDPO loses tool/action behavior during training, while the SGCD trace
    retains nonzero tool use and does not show the zero-tool fixed point. These training traces diagnose the
    training-time failure mode; Table~\ref{tab:tau3} reports the held-out
    mean Pass$^1$ comparison.}
  \label{fig:tau_training_diag}
\end{figure*}

\begin{figure*}[t]
  \centering
  \includegraphics[width=0.96\textwidth]{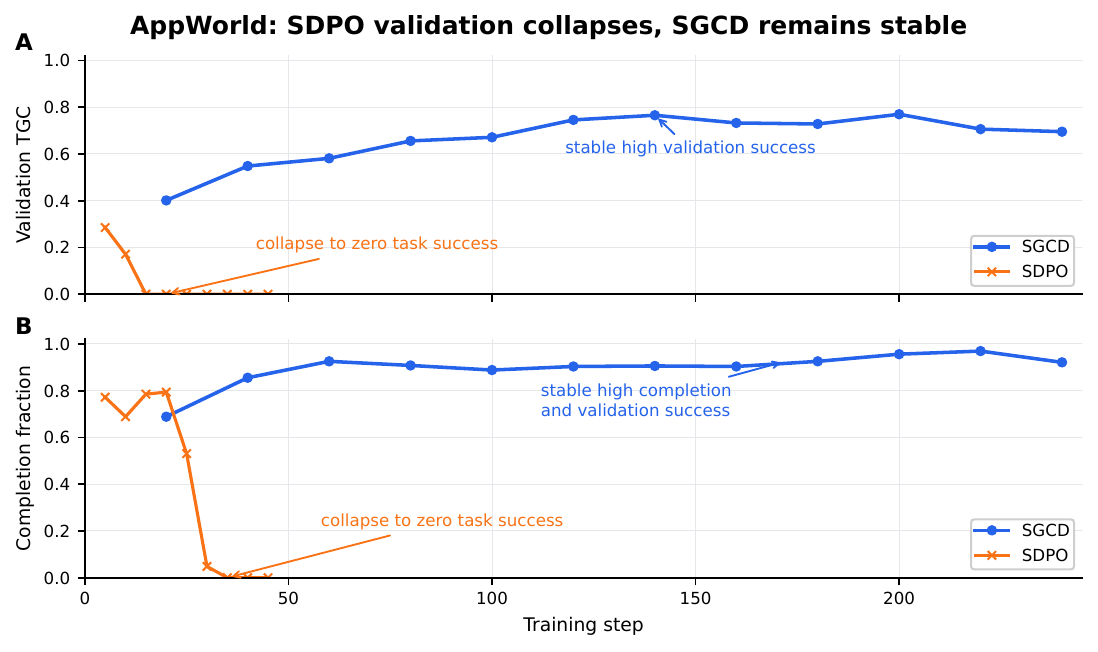}
  \caption{AppWorld training diagnostic trajectories. In the available diagnostic trace,
    SGCD maintains high validation and completion values through the 240-step run, while the SDPO diagnostic run
    collapses early in both task success and completion-related signals. The
    AppWorld SDPO curve stitches resumed diagnostic runs and is used as failure
    evidence, not as a replacement for the held-out result table. It is a
    separate training-trajectory diagnostic and is not numerically comparable
    to the held-out direct/uncapped ablation row in Table~\ref{tab:ablation}.}
  \label{fig:appworld_training_diag}
\end{figure*}

The held-out tables in the main paper are the primary result. Figures
\ref{fig:tau_training_diag} and \ref{fig:appworld_training_diag} provide
training-trajectory diagnostics: the $\tau^3$ plot tracks tool-use behavior,
while the AppWorld plot tracks validation success and completion.

\FloatBarrier

\section{AppWorld SFT Diagnostic}
\label{sec:sftdiag}

Table~\ref{tab:appworld_sft_diag} summarizes the diagnostic AppWorld SFT ladder
that motivated direct-from-base RL for AppWorld. These runs used an early
\texttt{batch\_001} diagnostic subset, so they are not numerically comparable to
the held-out \texttt{test\_normal}/\texttt{test\_challenge} headline table. They
nevertheless show the qualitative issue: standard SFT reduced next-token loss
while sharply degrading executable AppWorld behavior.

\begin{table}[h]
  \centering
  \venuecompacttablefont
  \setlength{\tabcolsep}{3pt}
  \begin{tabular}{lccc}
    \hline
    \textbf{Run} & \textbf{Rows / traj.} & \textbf{TGC} & \textbf{Test-pass} \\
    \hline
    Base 4B       & -- & \textbf{15.8} & \textbf{51.5} \\
    SFT v0        & 770 turn rows & 1.1 & 17.7 \\
    SFT v0.5      & 4,288 turn rows & 0.6 & 15.3 \\
    SFT v0.6      & 393 trajectories & looping & -- \\
    \hline
  \end{tabular}
  \caption{Diagnostic AppWorld SFT regression on a pooled
    \texttt{test\_normal}+\texttt{test\_challenge} \texttt{batch\_001}
    subset (177 trajectories: 51 normal and 126 challenge; 1,363 executable
    unit tests). TGC counts are 28/177, 2/177, and 1/177 for Base 4B, SFT v0,
    and SFT v0.5; corresponding test-pass counts are 702/1,363, 241/1,363, and
    208/1,363. The result
    motivates starting AppWorld SGCD from the base 4B model, unlike the
    $\tau^3$-airline ladder where SFT is a viable initialization.}
  \label{tab:appworld_sft_diag}
\end{table}

\end{document}